\documentclass[letterpaper, 10 pt, conference]{ieeeconf}  % Comment this line out if you need a4paper

\IEEEoverridecommandlockouts                              % This command is only needed if 
\usepackage{graphics} % for pdf, bitmapped graphics files
\usepackage{epsfig} % for postscript graphics files
\usepackage{times} % assumes new font selection scheme installed
\usepackage{amsmath} % assumes amsmath package installed
\usepackage{amssymb}  % assumes amsmath package installed
\usepackage{algorithm}
\usepackage{algorithmicx}
\usepackage{multirow}
\usepackage{graphicx}
\usepackage{subcaption}
\usepackage{algpseudocode}% http://ctan.org/pkg/algorithmicx
\usepackage{tikz}
\usepackage{csquotes}
\usepackage{bbm}
\usepackage{pgfplots}
\usepackage{xcolor}
\algnewcommand\RETURN{\State \algorithmicreturn}%

\algrenewcommand\algorithmicrequire{\textbf{Input:}}
\algrenewcommand\algorithmicensure{\textbf{Output:}}
\usepackage{cite}
\definecolor{rev}{RGB}{0,0,0}

\usepackage{tikz}
\usepackage{pgfplots}
\usepackage{xcolor}
\usepackage[normalem]{ulem}

\pgfplotsset{compat=1.18}
\usetikzlibrary{positioning}
\usetikzlibrary{decorations.pathreplacing}
\usetikzlibrary{calc}
\usetikzlibrary{positioning,shapes,shadows,arrows}

\pgfplotsset{
every axis/.append style={
  axis line style={->}, % arrows on the axis
  legend style={font=\scriptsize},
  label style={font=\scriptsize},
  title style={font=\scriptsize},
  tick label style={font=\scriptsize},
  }
}
\algblock{ParFor}{EndParFor}
\algnewcommand\algorithmicparfor{\textbf{parfor}}
\algnewcommand\algorithmicpardo{\textbf{do}}
\algnewcommand\algorithmicendparfor{\textbf{end\ parfor}}
\algrenewtext{ParFor}[1]{\algorithmicparfor\ #1\ \algorithmicpardo}
\algrenewtext{EndParFor}{\algorithmicendparfor}

\makeatletter
\let\NAT@parse\undefined
\makeatother
\usepackage{hyperref}

\title{\LARGE \bf
B4P: Simultaneous Grasp and Motion Planning for Object Placement \\via Parallelized Bidirectional Forests and Path Repair

\thanks{The authors are with the Department of Computer Science, Rice University, Houston, TX 77005, USA. This project is supported by the US National Science Foundation grant FRR-2240040.}
}

\author{Benjamin H. Leebron, Kejia Ren, Yiting Chen, and Kaiyu Hang}

\begin{document}

\maketitle
\thispagestyle{empty}
\pagestyle{empty}

\begin{abstract}
Robot pick and place systems have traditionally decoupled grasp, placement, and motion planning to build sequential optimization pipelines with an assumption that the individual components will be able to work together. However, this separation introduces sub-optimality, as grasp choices may limit, or even prohibit, feasible motions for a robot to reach the target placement pose, particularly in cluttered environments with narrow passages. To this end, we propose a forest-based planning framework to simultaneously find grasp configurations and feasible robot motions that explicitly satisfy downstream placement configurations paired with the selected grasps. Our proposed framework leverages a bidirectional sampling-based approach to build a start forest, rooted at the feasible grasp regions, and a goal forest, rooted at the feasible placement regions, to facilitate the search through randomly explored motions that
connect valid pairs of grasp and placement trees. We demonstrate that the framework’s inherent parallelism enables superlinear speedup, making it scalable for applications for redundant robot arms, e.g., 7 DoF, to work efficiently in highly cluttered environments. 
Extensive experiments in simulation demonstrate the robustness and efficiency of the proposed framework in comparison with multiple baselines under diverse scenarios.
\end{abstract}

\section{Introduction}
  Robot pick and place is a fundamental manipulation skill needed in various application scenarios \cite{correll2016analysis, wisspeintner2009robocup}. To relocate an object to the target pose, such systems are required to: 1) find a feasible grasp configuration; 2) find a feasible placement configuration; and 3) generate a feasible motion plan to connect theses two configurations under kinematic and environmental constraints. Traditionally, these topics have been investigated individually e.g., grasp planning \cite{morrison2018closing, fang2020graspnet, hang2016hierarchical}, motion planning \cite{Elbanhawi2014}, and placement planning \cite{Harada12}. More recently, they have been also studied as coupled problems, e.g., motion planning for grasping or placement, and grasping for placement \cite{haustein2019object, GraspPlanningHaustein}.
  
\begin{figure}[t]
    \centering
        \includegraphics[width=0.8\columnwidth]{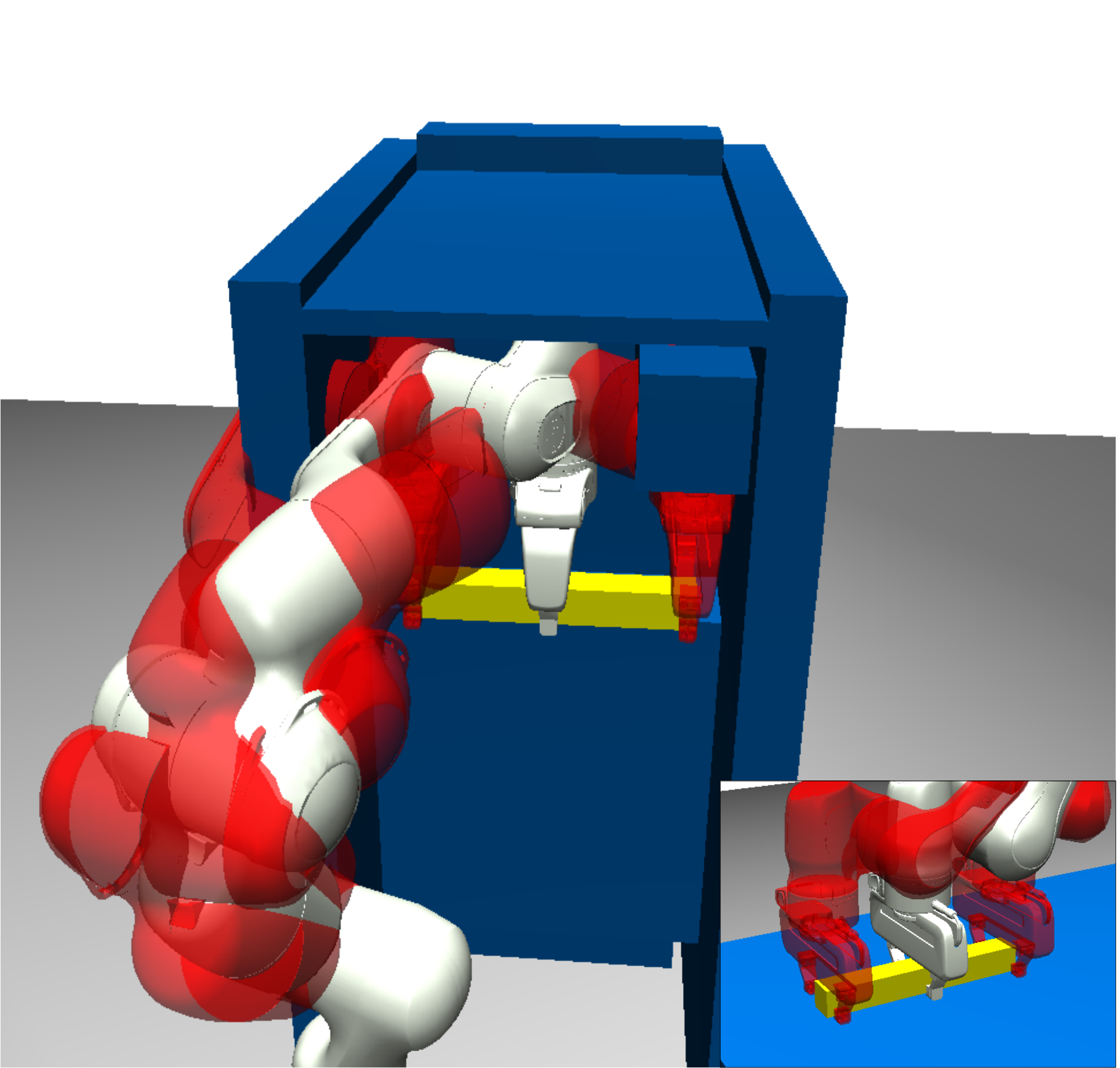}
    \caption{A motivational scenario where the robot is tasked to reposition the target object, the yellow stick, to the shelf. It must pass through a narrow passage and of the three pictured grasps, only the grasp in the center of the object (white) will allow the robot to reach the placement location. However, all grasps are feasible initially, and we must somehow find which initial grasp will allow for the robot to reach the downstream placement location.}
    \label{fig:motivation}
    \vspace{-0.5cm}
\end{figure}

  However, a unified framework that addresses pick, motion, and placement planning simultaneously to ensure that all internal constraints are satisfied is yet to be developed. For example, a planner needs to select a grasp that can pass through all narrow passages through a robot motion to finally reach a selected placement pose. Note that even if both the grasp and placement poses are valid, it is often an issue that there is no robot motion to connect them due to the collisions rendered by the selected grasp. See Fig.~\ref{fig:motivation} for an example.

% However, grasp pose selection beyond force closure alone is crucial for successful downstream motion planning, as the chosen grasp configuration serves as the starting state for computing a feasible trajectory to complete the placement task. Especially when the environment is cluttered or spatially constrained, the selected grasp pose needs to simultaneously satisfy a suitable placement configuration conjunct with a feasible robot trajectory plan. Achieving this dual requirement poses a significant challenge --- as a poorly chosen grasp can render even geometrically valid placement configurations unreachable due to joint limits and environmental constraints, it is essential to consider grasp and motion planning jointly.

To this end, we propose \textbf{B4P} (Bidirectionally Picking, Planning, and Placing in Parallel), a joint planning framework for object pick and place in spatially constrained environments. Specifically, the framework first identifies a grasp region -- a set of robot configurations that ensure a feasible grasp; and a placement region -- a set of configurations that reach the target placement poses. We do not make any assumptions on how such regions can be found so that any existing planner can be used, e.g., \cite{hang2016hierarchical}. Thereafter, through a sampling-based scheme, the proposed B4P builds a \emph{pick forest} (with trees rooted inside the identified grasp region) and a \emph{place forest} (with trees rooted inside the identified placement region), and simultaneously grows those trees bidirectionally to find paired connections between starts and goals. Once a connection is made, a grasp configuration and a placement configuration will be paired, and a complete robot motion will be simultaneously provided.

% Subsequently, it connects these regions using a bidirectional configuration space forest, ensuring smooth and feasible transitions from grasping to placement even in spatially constrained environments.

Our B4P algorithm design offers an inherent internal structure that can be parallelized to achieve superlinear speedups thanks to its non-deterministic nature of sampling-based algorithms \cite{plaku2005sampling}. The framework is also designed to offer application flexiblity, allowing modular integration with any grasp method that can find feasible end-effector poses, and with any placement planning methods that can generate end-effctor or object final poses. In brief, the proposed B4P:
\begin{enumerate}
    \item finds grasps that are guaranteed to be suitable for downstream placement through narrow passages or in cluttered environments;
    \item leverages the parallelizability of forest-based planning and to achieve a superlinear speedup;
    \item can smoothly integrate with any grasp planners and placement planners to work with various robots. 
\end{enumerate}

\section{Related Works}

\subsubsection{Grasp and Placement Planning}
Grasp planning is a critical component in robot pick and place systems, focusing on determining how the robot's end-effector should be placed to pick up an object securely. Current research on grasp planning can be roughly divided into two categories: model-based~\cite{bicchi2000robotic} and data-driven~\cite{newbury2023deep}. Model-based methods~\cite{hang2016hierarchical, ciocarlie2009hand, liu2021synthesizing, chen2023differentiable} rely on prior knowledge of the target object and analytically compute grasp configurations that meet grasp criteria such as force closure. Data-driven approaches~\cite{sundermeyer2021contact, fang2020graspnet, morrison2018closing} aim to derive a direct mapping from visual input to grasp poses from well-annotated datasets. On the other end of the problem spectrum, placement planning has investigated object-centric optimization problems, such as placement stability and functionality, as well as task and environmental constraints \cite{haustein2019object}. However, it is still an open problem in the field that grasp and placement configurations, when planned separately, are not necessarily connectable with any feasible robot motions.

% To advance robotic pick and place systems with state-of-the-art research, the proposed framework allows modular integration of any grasp planning approaches. Given a set of feasible end-effector poses, the proposed framework ensures a grasp pose selection compatible with the downstream placement task. 

\subsubsection{Sampling-based Motion Planning}
Motion planning computes a collision-free path in the robot's configuration space to connect the grasp with the placement configuration in pick and place systems. Sampling-based planners~\cite{orthey2023sampling} have proven efficient and generalizable in exploring the high-dimensional configuration space of manipulators. Significant acceleration can be obtained through parallelism~\cite{thomason2024motions, bialkowski2011massively, fishman2023motion, sundaralingam2023curobo, bhardwaj2022storm} on both GPU and CPU-based computing. Key properties of sampling-based planners include probabilistic completeness~\cite{kuffner2000rrt} and Voronoi bias~\cite{lindemann2004incrementally}. The former ensures that a solution will eventually be found as long as it exists in the search space, while the latter
contributes to exploration efficiency in high-dimensional spaces. Based on the advantages of sampling-based approaches, and along with a novel forest-based planning strategy, the proposed B4P can simultaneously pair valid grasp and placement configurations while generating feasible motion plans to connect them.

% Therefore, the proposed framework employs a sampling-based bidirectional forest to balance completeness and efficiency in joint grasp pose selection and placement motion planning.

\subsubsection{Pick and Place Planning}
Pick and place planning in robotics primarily focuses on grasping a target object and relocating it to a specified location. Grasp planning and motion planning are both critical components during this process. Zeng~\cite{zeng2022robotic} presents a system design solely focusing on multi-modality grasp planning in the Amazon Robotics Challenge. Saut~\cite{saut2010planning} proposes a dual-arm approach to achieve larger workspace. Haustein~\cite{haustein2019object} aims at finding a feasible placement configuration in cluttered environments. While taking placement compatibility into consideration during grasp planning~\cite{shanthi2024pick, he2023pick2place}, however, a geometrically valid placement configuration might still be unreachable~\cite{brooks1983planning} by any collision-free robot motions due to kinematic and environmental constraints. In an effort to unify and integrate these key functionalities, our B4P focuses on building a planning that can be compatible with any existing pick planning and placement planning approaches, while coordinate in the middle to ensure the compatibility of individual solutions.

\section{Preliminaries and Problem Statement}
\label{prob_formulation}
In this section, we first review the preliminaries of the proposed work, based on which we formulate robot pick and place as a problem of simultaneously finding pick, motion, and placement solutions.

\subsection{Preliminaries}

In this work, we consider the pick and place problem as relocating a target object from a start pose to a target pose through a collision-free robot motion. To unify our planning framework as aforementioned, the grasp configurations, the placement configurations, as well as the robot motion paths are all expressed in the \emph{robot configuration space}.

The configuration space of a $n-$DoF robot is denoted by $\mathcal{Q}\subset \mathbb{R}^n$. The robot's forward kinematics is denoted by $\Gamma: \mathcal{Q} \mapsto SE(3)$, such that any valid robot joint configuration $q \in \mathcal{Q}$ can be mapped to an robot end-effector's pose $x \in SE(3)$. Accordingly, the inverse kinematics is denoted by $\Gamma^{-1}$. Note that these definitions are not limited to arm robots, but also can be applied to any robots that have a controlled end-effector. For example, as shown later in Fig.~\ref{fig:repair_example}, a mobile robot with a fixed gripper can also map from its base configuration to an end-effector's pose.

For an object inside the workspace of the robot, let us denote its pose by $x^{obj} \in SE(3)$, a grasp planner will be able to generate feasible pick configuration relative to $p^{obj}$. Let us denote by $\Call{PlanPick}{\cdot}: SE(3) \mapsto \mathcal{Q}$ the function of this planner that generates picking configurations:
\begin{equation}
    q^{pick} = \Call{PlanPick}{x^{obj}}
    \label{eq:plan_pick}
\end{equation}
In practice and for many grasp planners, every given $p^{obj}$ can result in $0, 1$ or many picking configuration solutions. Without loss of generality, in this work as we focus on building the joint framework for pick, motion, and place planning that is compatible with existing grasp planners, we treat $\Call{PlanPick}{\cdot}$ as a sampler that can generate a picking configuration every time it is called. The output configuration can be the same or different across function calls.

Inside the workspace of the robot, let us denote by a region $Y \subset SE(3)$ where the placement of an object is expected. A placement planner, which finds a pose within $Y$ for the object, is denoted as $\Call{PlanPlace}{\cdot}: SE(3) \mapsto SE(3)$ to generate a stable placement under given task constraints:
\begin{equation}
    x^{place} = \Call{PlanPlace}{Y}
    \label{eq:plan_place}
\end{equation}
Similarly to grasp planning, to be compatible with exisiting placement planners, we treat this place planner as a sampler that can generate the same, or different, placement plans every time it is called. Furthermore, to keep our planning framework operating with robot configurations only as aforementioned, every planned placement pose will be converted to a robot configuration by:
\begin{equation}
    q^{place} = \Gamma^{-1}(x^{place}(x^{obj})^{-1}\Gamma(q^{pick}))
    \label{eq:q_place}
\end{equation}
where the end-effector's pose at placement calculated by $x^{place}(x^{obj})^{-1}\Gamma(q^{pick})$ ensures that the grasp on the object has not changed during the robot motion execution.

\subsection{Trees and Forests}

Same as in most sampling-based motion planning approaches, we use trees to represent the exploration of the valid robot configurations and robot motions to connect between them in the configuration space $\mathcal{Q}$. In our bidirectional framework, we denote by $\mathcal{T}^{pick}_i$ a motion tree rooted at the $i-$th sampled picking configuration, and by $\mathcal{T}^{place}_j$ a motion tree rooted at the $j-$th sampled placement configuration. \textcolor{rev}{We also denote $\mathcal{T}^{pick}_i.pick$ to be the $SE(3)$ transformation of the gripper relative to the object.} In the trees, every node is a collision-free robot configuration and every edge is a collision free path. To enable all trees to simultaneously explore the configuration space and find potential valid pairings between $\mathcal{T}^{pick}_i$ and $\mathcal{T}^{place}_j$, we collect the trees into two motion forests $\mathcal{F}^{pick} = \{\mathcal{T}^{pick}_i\}_{i=1:n}$ and $\mathcal{F}^{place} = \{\mathcal{T}^{place}_j\}_{j=1:m}$, with $n$ and $m$ trees respectively. While the trees grow from two forests toward each other, only connections between trees from different forests are allowed.

\subsection{Problem Formulation}

The pick, motion, and place planning problem in this work is formalized as follows. Given an object in the robot workspace at $x^{obj}$, and a placement region $Y$, find a picking configuration $q^{pick}$ for $x^{obj}$, and a robot motion plan $\pi = \{q^{pick}, \ldots, q_k, \ldots, q^{place}\}$, such that:
\begin{itemize} 
    \item the final object pose $x^{place}$, as calculated by Eq.~\ref{eq:q_place}, satisfies $ x^{place} \in Y$;
    \item all intermidiate robot states $q_k$ together with the object with its in-hand pose determined by $q^{pick}$ at the beginning will be collision-free.
\end{itemize}

% \begin{algorithm}
% \caption{B4P}\label{alg:cap}
% \begin{algorithmic}[1]
% \renewcommand{\algorithmicrequire}{\textbf{Input:}}
% \renewcommand{\algorithmicensure}{\textbf{Output:}}
% \Require Start region $Q_S$, Goal region $Q_G$, Grasp sampler $GS$
% \State $F \leftarrow {MakeStarts}(Q_S, GS)$
% \State $B \leftarrow {MakeEnds}(Q_G)$
% \While{Time Remaining}
% \While{$AwaitConnection()$}
% \State{$ConnectGraph(F_S, F_G)$} \COMMENT{Alg 2}
% \EndWhile
% \State{$\pi' \leftarrow PathRepair(connection)$} \Comment{Alg 3}
% \If{$\pi'$ is valid}
% \Return{ $\pi'$}
% \ElsIf
% \State {$ResetConnection()$}
% \EndIf
% \EndWhile
% \end{algorithmic}
% \end{algorithm}
% We formulate the problem as follows: Given a set of initial configurations $\pi(0) \in Q_S$ and a desired final configuration set $Q_G$, we want to find a time $T$, a grasp $\theta \in \Theta$ and a set of waypoints $\pi : [0, T] \rightarrow Q_{free}$ such that the motion satisfies $\pi(T)\in Q_{G}$.
% We formulate the problem as: Given a $\Theta_s$ and $\Theta_g$ in the Cartesian workspace, we aim to identify both $S\in \mathcal{Q}$ and $G\in\mathcal{Q}$ and find the path $\pi$ that connects $S$ and $G$ in the collision-free configuration space $\mathcal{Q}_{free}$. 

\section{Method: B4P}

We begin this section with an overview of the proposed B4P algorithm and then delve into the details of its key components of forest building and path repair.

\begin{algorithm}[t]
\caption{The B4P algorithm}
% \small
\footnotesize
    \begin{algorithmic}[1]
        \Require Object pose $x^{obj}$, placement region $Y$, number of pick and place trees $N^{pick}$ and $N^{place}$
        \Ensure Robot motion path $\pi$
        \State $\mathcal{F}^{pick}, \mathcal{F}^{place} \gets \Call{SpawnForest}{x^{obj}, Y, N^{pick}, N^{place}}$
        \Comment{Alg.~\ref{alg:spawn_forest}}
        \State Workers $\gets$ \Call{LaunchParaWorkers}{$\mathcal{F}^{pick}, \mathcal{F}^{place}$}
        \ParFor{w $\in$ Workers}
        \Comment{Parallel Workers}
            \If{w.\Call{BuildForest}{\null}}
            \Comment{Alg.~\ref{alg:worker}}
                \State $\pi \gets $ w.\Call{Path}{\null}
                \Comment{Initial Path}
                \State $\pi \gets$ \Call{PathRepair}{$\pi$}
                \Comment{Alg.~\ref{alg:path_repair}}
                    \If{$\pi \ne \{\}$}
                    \Comment{Success}
                    \State Workers.\Call{Finish}{\null}
                    \EndIf
            \EndIf
        \EndParFor

        \State \Return $\pi$
    \end{algorithmic}
    \label{alg:B4P}
\end{algorithm}

\subsection{Algorithm Overview}
\label{B4P_overview}

In Eq.\eqref{eq:q_place} we can see that a robot configuration $q^{place}$ for placement can be calculated only if the initial picking configuration $q^{pick}$ is known. While exact pick and place configuration pairing is ensured, this requirement, however, will pose a hard constraint that a pick tree $\mathcal{T}^{pick}_i$ and a place tree $\mathcal{T}^{place}_j$ in the forests can be potentially connected only if their roots share the same picking configuration relative to the object, i.e., $\mathcal{T}^{pick}_i.pick = \mathcal{T}^{place}_j.pick$. As such, the bidirectional forests will be divided by root configurations into sub-forests, and the ability of fully exploring motion possibilities will be significantly reduced.

\begin{algorithm}[t]
\caption{SpawnForest($\cdot$)}
% \small
\footnotesize
    \begin{algorithmic}[1]
        \Require Object pose $x^{obj}$, placement region $Y$, number of pick and place trees $N^{pick}$ and $N^{place}$
        \Ensure Pick forest $\mathcal{F}^{pick}$, place forest $\mathcal{F}^{place}$
        \State $\mathcal{F}^{pick} \gets \{\}, \mathcal{F}^{place} \gets \{\}$
        \For{$i = 1, \ldots, N^{pick}$}
            \State $q^{pick}$ = \Call{PlanPick}{$x^{obj}$}
            \Comment{Eq.~\eqref{eq:plan_pick}}
            \State $\mathcal{F}^{pick}.\Call{AddRoot}{q^{pick}}$
        \EndFor
        \For{$i = 1, \ldots, N^{place}$}
            \State $x^{place}$ = \Call{PlanPlace}{$Y$}
            \Comment{Eq.~\eqref{eq:plan_place}}
            \State $q^{pick}$ = \Call{PlanPick}{$x^{obj}$}
            \State $q^{place} = \Gamma^{-1}(x^{place}(x^{obj})^{-1}\Gamma(q^{pick}))$
            \Comment{Eq.~\eqref{eq:q_place}}
            \State $\mathcal{F}^{place}.\Call{AddRoot}{q^{place}}$
        \EndFor

        \State \Return $\mathcal{F}^{pick}, \mathcal{F}^{place}$
    \end{algorithmic}
    \label{alg:spawn_forest}
\end{algorithm}

\begin{algorithm}[t]
\caption{Worker.BuildForest()}
\footnotesize
    \begin{algorithmic}[1]
        \Require Worker w
        \Ensure Boolean done
        \State done $\gets false$
        \While{$\Call{Time.Available}{\null}$ $\land$ \textbf{not} done}
            \State $q \gets$ \Call{SampleRobotConfig}{\null}
            \State $\mathcal{T}^{pick}_{near} \gets$ \Call{FindNearestPickTree}{$q$}
            \State $q_{new}^{pick} \gets$ \Call{ExpandTree}{$\mathcal{T}^{pick}_{near}, q$}
            \Comment{Forward Expansion}
            \Statex
            \State $\mathcal{T}^{place}_{near} \gets$ \Call{FindNearestPlaceTree}{$q$}
            \State $q_{new}^{place} \gets$ \Call{ExpandTree}{$\mathcal{T}^{place}_{near}, q$}
            \Comment{Backward Expansion}
            \Statex
            \If{\Call{Connect}{$q_{new}^{pick}, q_{new}^{place}$}}
            \Comment{Try Pairing}
                \State Worker.Path $\gets$ \Call{ExtractPath}{$\mathcal{T}^{pick}_{near}, \mathcal{T}^{place}_{near}$}
                \State done $\gets$ true
            \EndIf
        \EndWhile        
        \State \Return \textcolor{rev}{done}
    \end{algorithmic}
    \label{alg:worker}
\end{algorithm}

To this end, in our work, as outlined in Alg.~\ref{alg:B4P}, in order to facilitate the tree expansion by sufficiently exploring pairings between the two forests, we opt to omit the constraints of an explicit picking configuration for all place trees $\mathcal{T}^{place}_j$. For this, when a placement pose $x^{place}$ is sampled, a random picking configuration will be assigned to it to compute $q^{place}$. When our algorithm B4P expands the forests from both sides and tries to make connections, the condition of $\mathcal{T}^{pick}_i.pick = \mathcal{T}^{place}_j.pick$ is not checked when making connections between pick and place trees. Once a connection is made, it is possible that the picking configuration $q^{pick}$ of the pick tree's root does not match with the the $q^{place}$ in the place tree's root per the constraints in Eq.~\eqref{eq:q_place}. In that case, B4P recalculates $q^{place}$ for the place tree, using Eq.~\eqref{eq:q_place} and the $q^{pick}$ from the pick tree, to enforce the pairing to create an initial path (line \#5 in Alg.~\ref{alg:B4P}). 

However, when initially the pick and place trees used different picking configurations, the collision checking for them was also using different object poses in the hands. After enforcing the place tree to take the picking configuration from the paired pick tree, obtained trajectory corresponding to the place tree can potentially have collisions. To address this problem, our proposed B4P develops a post-planning \emph{path repair} mechanism (line \#6 in Alg.~\ref{alg:B4P}), as detailed in Sec.~\ref{sec:path_repair}, to locally fix minor collisions in an efficient way.

\begin{figure}[t]
    \centering
        \includegraphics[width=0.7\columnwidth]{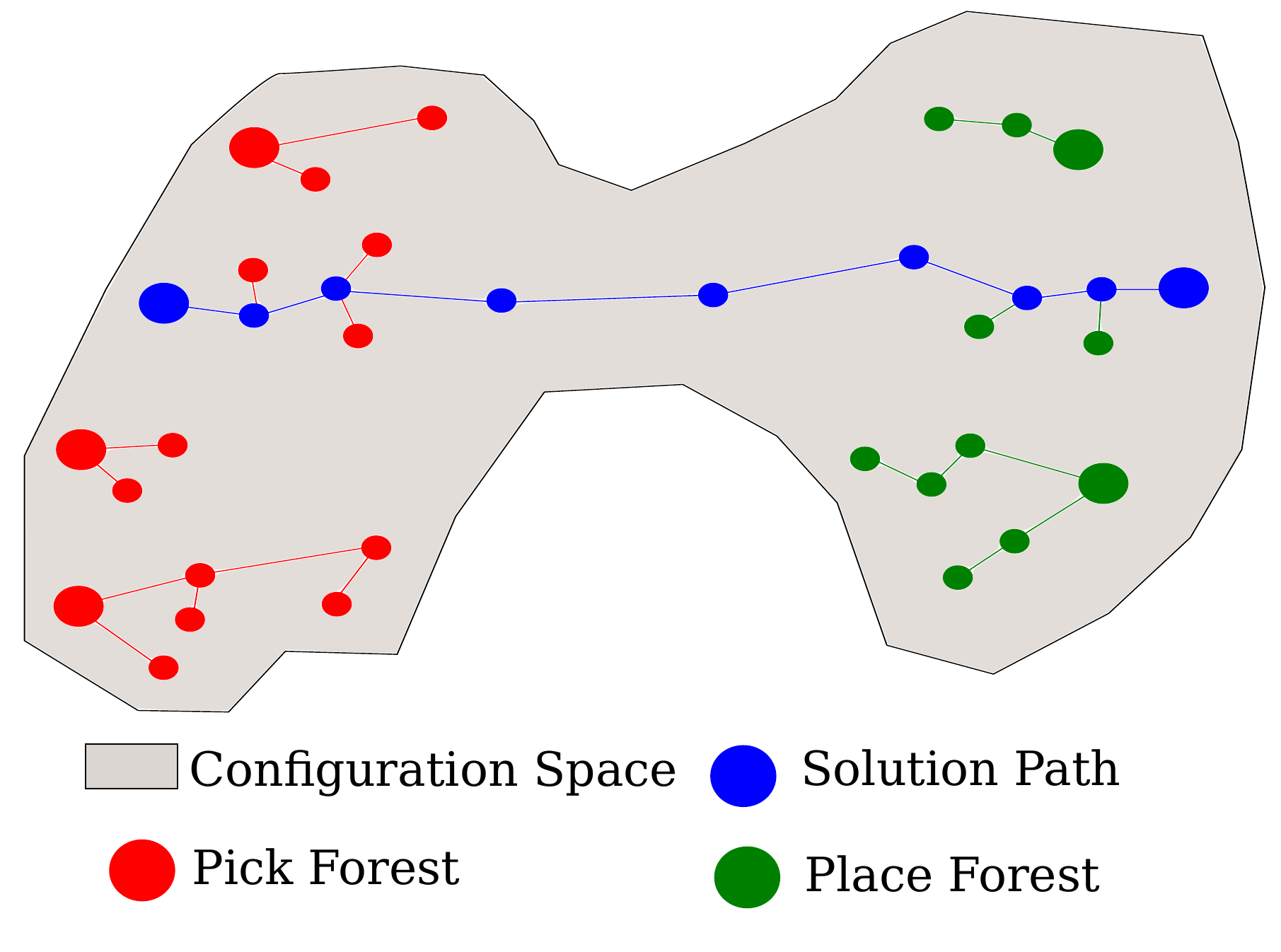}
    \caption{An illustration of the forest building in the robot configuration space with $4$ pick trees and $3$ place trees.}
    \label{fig:forest_example}
\end{figure}

\begin{figure}[t]
    \centering
        \includegraphics[width=0.9\columnwidth]{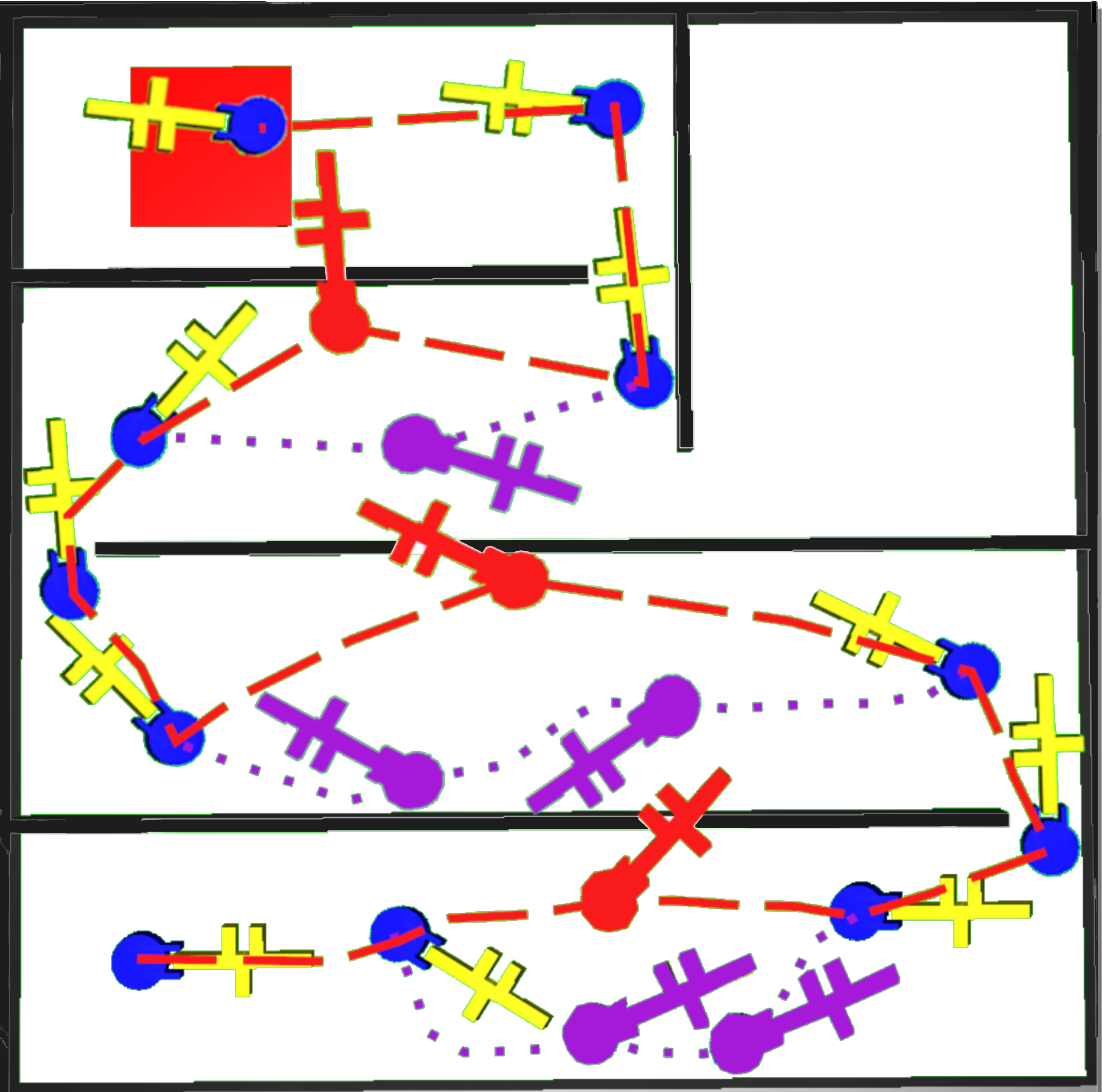}
    \caption{An example path repair procedure for a 2D mobile robot (blue) pick and place task. The red dashedline represents the initial path before repair, and the purple dotted lines are the repairs made. States in red are collisions when the grasp is used, and states in purple are the replanned nodes. }
    \label{fig:repair_example}
    \vspace{-0.4cm}
\end{figure}

\begin{figure*}[h]
    \centering
    \footnotesize
    % \begin{tikzpicture}
    %     \node[anchor=south west, inner sep=0] at (0,0){\includegraphics[width=2\columnwidth]{Plots_and_Plans/2x3Collage.pdf}};
    %     \node[anchor=west, align=left] at (0.3, 4){A. 2D Maze};
    %     \node[anchor=west, align=left] at (3.55, 4.8){\textcolor{green}{B. Stick on Shelf}};
    %     \node[anchor=north east, align = left] at (8.1, 5){\textcolor{white}{C. Grocery-shelf}};
    % \end{tikzpicture}
        \includegraphics[width=2\columnwidth]{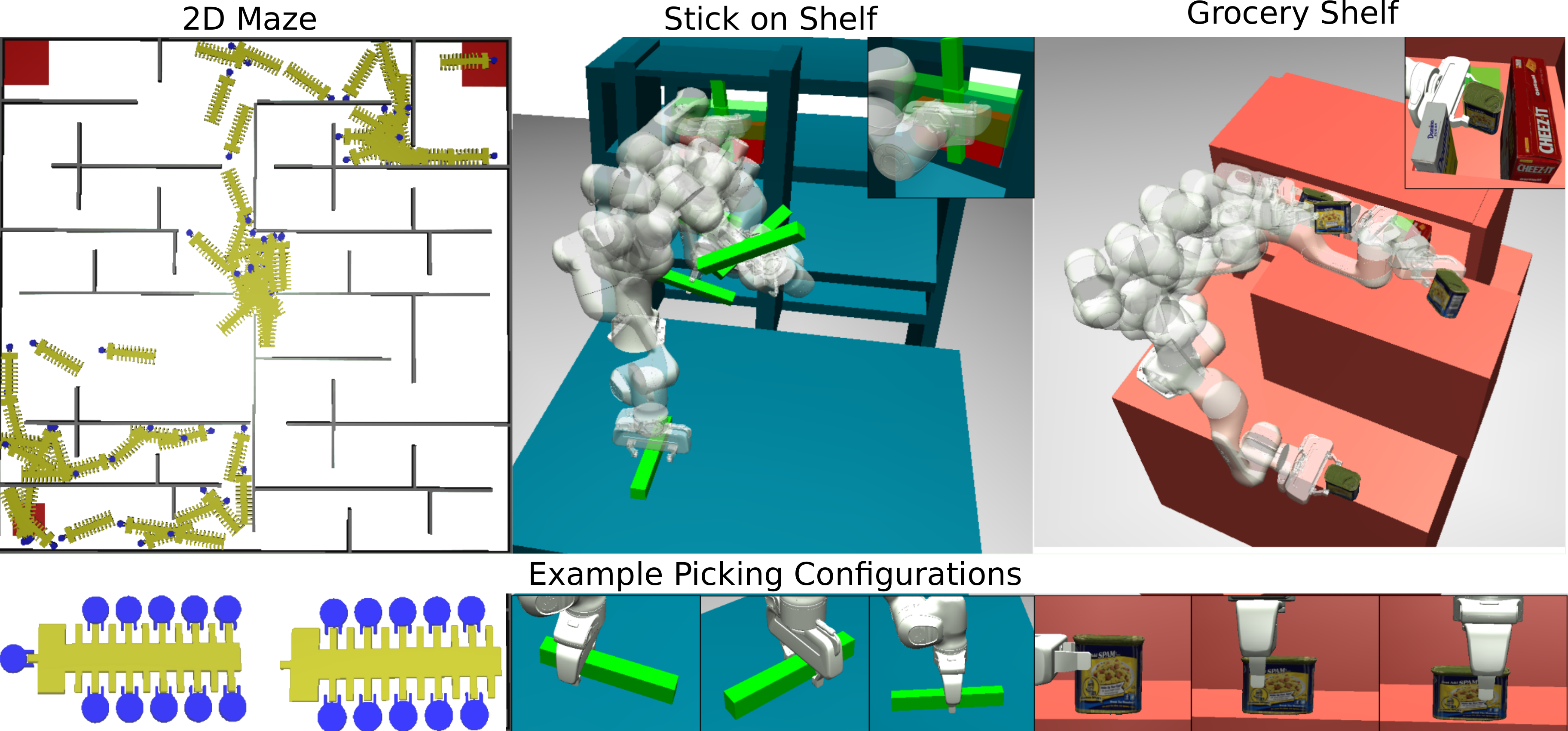}
    \caption{[2D Maze] The robot needs to find a feasible path to carry the grasped object from the start pose (left bottom corner) to the target region (red area in the right top corner); [Stick on Shelf] The manipulator needs to grasp the stick and place it on the target region on shelf; [Grocery Shelf] The manipulator needs to grasp the target object on the shelf in a grocery scenarios; [Example Picking Configurations] Left: Sampled grasp poses on each edge of the planar object; Middle: Sampled grasp poses on the green stick; Right: Sampled grasp poses on the spam object. \textcolor{rev}{For each task, grasp poses are uniformly randomly sampled. The optimal number of pick and place trees generated depends on the task.}} 
    \label{fig:environments}
\label{fig:exps_tasks}
\vspace{-0.4cm}
\end{figure*}
\subsection{Path Planning by Building Forest}

B4P begins by spawning a forest with trees rooted at both pick and place configuration regions. For this, as detailed in Alg.~\ref{alg:spawn_forest}, the forests are initialized as empty sets, and then iteratively populated with roots provided by $\Call{PlanPick}{\cdot}$ and $\Call{PlanPlace}{\cdot}$. Once the roots are ready, B4P launches parallel workers to grow the trees from both pick and place forests toward each other. As described in Alg.~\ref{alg:worker}, each worker is independently working on building the shared forest. By sampling a random robot configuration $q \in \mathcal{Q}$ (line \#3), B4P first finds the nearest tree $\mathcal{T}^{pick}_{near}$ in the pick forest $\mathcal{F}^{pick}$ that has a node closest to $q$. The tree $\mathcal{T}^{pick}_{near}$ will then expand towards $q$ with a linear motion as much as possible, until collisions are detected, to add a new configuration $q_{new}^{pick}$ into $\mathcal{T}^{pick}_{near}$. In the backward direction, this worker will then try to grow the nearest place tree $\mathcal{T}^{place}_{near}$ towards $q$ to add a new configuration $q_{new}^{place}$.

An important step in Alg.~\ref{alg:worker} is the active attempt to connect $q_{new}^{pick}$ and $q_{new}^{place}$ by every worker in every iteration (line \#8). If such a connection can be made with a valid motion, an initial solution path will be constructed. If not, the worker will continue to build the forest until one of the workers has found a solution. A configuration space illustration of Alg.~\ref{alg:worker} is shown in Fig.~\ref{fig:forest_example}. Note that dependent on the sampled configuration $q$, a parallel worker can work on different pairs of pick and place trees in different iterations, so that the work spent on different trees are fully determined by the random samples and not biased by any root picking or placement configurations.

\begin{algorithm}[t]
\caption{PathRepair($\cdot$)}
\footnotesize
    \begin{algorithmic}[1]
        \Require Motion path $\pi$
        \Ensure Repaired motion path $\pi^*$
        \State $\pi^* \gets \{\pi.\Call{GetNote}{1}\}$
        \State collision $\gets false$
        \State $k \gets 0$
        \For{$i = 1, \ldots, \pi.\Call{Len}{\null}-1$}
            % \For{$j = 2, \ldots, \pi.\Call{Len}{\null}$}

                \State begin $\gets \pi.\Call{GetNode}{i-k}$
                \State temp $\gets \pi.\Call{GetNode}{i}$
                \State end $\gets \pi.\Call{GetNode}{i+1}$
                
                \If{\Call{CheckCollision}{temp, end}}
                \Comment{Collision Found}
                    \State $k \gets k+1$
                    \State collision $\gets true$
                \ElsIf{collision $\land$ \textbf{not} \Call{CheckCollision}{temp, end}}
                    \State $\eta \gets$ \Call{ParaLocalReplan}{begin, end}
                    \Comment{Locally Repair}
                    \If{$\eta \ne \{\}$}
                        \State $\pi^*$.\Call{AddPath}{$\eta$}
                        \State $k \gets 0$
                    \Else
                        \State \Return $\{\}$
                        \Comment{Repair Failed}
                    \EndIf
                \Else
                    \State $\pi^*$.\Call{AddNode}{end}
                    \Comment{Collision-Free Nodes}
                \EndIf        
        \EndFor        
        \State \Return $\pi^*$
        \Comment{Success}
        
    \end{algorithmic}
    \label{alg:path_repair}
\end{algorithm}

\subsection{Path Repair}
\label{sec:path_repair}

As discussed above, an initial solution found by Alg.~\ref{alg:worker} can contain local collisions due to the enforced pairing of pick and place trees. As shown in line \#5-6 of Alg.~\ref{alg:B4P}, such an initial solution will go through a $\Call{PathRepair}{\cdot}$ procedure to eliminate such collisions to produce a completely collision-free path for the pick and place task. Concretely, given an initial path $\pi$, $\Call{PathRepair}{\cdot}$ will iteratively check through every waypoint in $\pi$ and find all continuous collision-involved segments to repair. As detailed in Alg.~\ref{alg:path_repair}, two pointers \emph{begin} and \emph{end} are used to keep track of the range of the collision-involved segments as the iterator $i$ moves through $\pi$. Once a segment is identified, meaning that there is no collision at the waypoints immediately before and after the segment, a fast parallel local repair procedure $\Call{ParaLocalReplan}{begin, end}$ will be invoked to generate a detour path $\eta$ to avoid the detected collisions. This $\Call{ParaLocalReplan}{begin, end}$ is a highly parallelized local motion planner implemented in a similar way to Alg.~\ref{alg:worker}, with the only modification that all workers are now working on a fixed pair of \emph{begin} and \emph{end} roots.

% \begin{figure*}[h]
%     \centering
%     \footnotesize
%     % \begin{tikzpicture}
%     %     \node[anchor=south west, inner sep=0] at (0,0){\includegraphics[width=2\columnwidth]{Plots_and_Plans/2x3Collage.pdf}};
%     %     \node[anchor=west, align=left] at (0.3, 4){A. 2D Maze};
%     %     \node[anchor=west, align=left] at (3.55, 4.8){\textcolor{green}{B. Stick on Shelf}};
%     %     \node[anchor=north east, align = left] at (8.1, 5){\textcolor{white}{C. Grocery-shelf}};
%     % \end{tikzpicture}
%         \includegraphics[width=2\columnwidth]{Plots_and_Plans/example_results.pdf}
%     \caption{[2D Maze] The robot needs to find a feasible path to carry the grasped object from the start pose (left bottom corner) to the target region (red area in the right top corner); [Stick on Shelf] The manipulator needs to grasp the stick and place it on the target region on shelf; [Grocery Shelf] The manipulator needs to grasp the target object on the shelf in a grocery scenarios; [Example Picking Configurations] Left: Sampled grasp poses on each edge of the planar object; Middle: Sampled grasp poses on the green stick; Right: Sampled grasp poses on the spam object.} 
%     \label{fig:environments}
% \label{fig:exps_tasks}
% \end{figure*}
When $\Call{ParaLocalReplan}{begin, end}$ finishes, it is possible that there is no path repair solution can be found. In that case, Alg.~\ref{alg:path_repair} will return an empty path to let B4P know that the search needs to continue. Otherwise, B4P will return with a successfully found path $\pi$ for the pick and place task (line \#7 in Alg.~\ref{alg:B4P}). An example path repair is visualized for a 2D mobile robot pick and place task in Fig.~\ref{fig:repair_example}.

\section{Experiments}
In this section, we first provide an overview of our experiments, including task design, baseline selection, and system environment. Then, we demonstrate the experimental result in both challenging 2D and 3D tasks to validate the effectiveness and efficiency of the proposed B4P framework. 

\subsection{Overview}
\subsubsection{Task Scenarios}

We evaluate the proposed framework across challenging 2D and 3D scenarios as illustrated in Figure.~\ref{fig:exps_tasks}, including:
\begin{itemize}
    \item a 2D maze task where a robot needs to pick the target object and navigate to the target region;
    \item a 3D stick-on-shelf task where a 7-DoF Franka Emika Panda manipulator needs to pick the target stick and safely place it in a specific region on the shelf;
    \item a 3D grocery-shelf task where the same manipulator needs to pick the target object and place it in a specific region in challenging grocery scenarios. 
\end{itemize} 
All tasks are designed to create a spatially constrained workspace for the robot; therefore, a majority of feasible grasp poses are incompatible with the task placement requirement due to infeasible placement configuration or unreachable motion trajectory plan. 

\subsubsection{Baselines}
To provide a comprehensive evaluation, we introduce two bidirectional RRT parallel algorithms as our baseline to compare with the proposed B4P approach. \textcolor{rev}{These planners sample 1 placement with their sampled grasp and only use one placement as a goal.} 
\begin{itemize}
    \item RRT-\textit{Individual}: Each thread \textcolor{rev}{$t$} performs bidirectional RRT using its own individual sampled grasp \textcolor{rev}{$q^{pick}_t$}. 
    \item RRT-\textit{Shared}: All threads perform \textcolor{rev}{one} bidirectional RRT \textcolor{rev}{in parallel} using a single grasp. In this approach, we set a small time limit and when it expires, the planner restarts and uses a new grasp. 
\end{itemize}

\subsubsection{System Environment}
All experiments are carried out on AMD Ryzen 9 5950X 16-Core Processor and 32GB of RAM on Ubuntu 20.04. We implemented the algorithms in C++ and adopted MuJoCo~\cite{mujoco} as the task simulator.

\subsection{2D Evaluation with Maze Task}
% \begin{figure}[h!]
%     \centering
%     \includegraphics[width=0.6\linewidth]{Plots_and_Plans/allgrasps.pdf}
%     \caption{All 21 grasps of the target object for the 2D Maze. Only the leftmost grasp in the top image is able to kinematically solve the maze.}
%     \label{fig:2dMazeEnv}
% \end{figure}
\begin{table}[h]
\centering
\setlength{\tabcolsep}{4pt} % Default value: 6pt

\begin{tabular}{|c|c|c|c|c|c|c|}
\hline
\#Threads & \multicolumn{2}{c|}{Shared} & \multicolumn{2}{c|}{Individual} & \multicolumn{2}{c|}{B4P (Ours)} \\
 & Time & Rate & Time & Rate & Time & Rate \\
\hline
1 & - & 0/10 & - & 0/10 & N/A & 0/10 \\
2 & - & 0/10 & - & 0/10 & 135.37 $\pm$ 32.56 & 3/10 \\
4 & - & 0/10 & - & 0/10 & 60.26 $\pm$ 25.9 & 10/10 \\
8 & - & 0/10 & - & 0/10 & 27.53 $\pm$ 14.81 & 10/10 \\
16 & - & 0/10 & 89.13 & 2/10 & 12.56 $\pm$6.66 & 10/10 \\
30 & - & 0/10 & 63.16 $\pm$ 37.69 & 4/10 & 5.42 $\pm$ 2.79 & 10/10 \\
\hline
\end{tabular}
\caption{Computation time (in seconds) for each method on the 2D Maze Task to output a feasible trajectory plan. The time budget is 120 seconds. 
\label{tab:2dmaze}
}
\end{table}

A 2D maze environment is designed to create a narrow passage in the plane for a robot to carry the target object to a target region. As shown in the bottom row of Fig.~\ref{fig:exps_tasks}, we consider each edge of the target object associated with one feasible grasp. Incompatible grasp poses will easily lead to getting stuck at some flexural corner in the maze. After a stable grasp is formed, we consider the object and the robot undergoes the same rigid body transformation. We evaluate our planner on success rate, time, and speedup for different numbers of threads and the result is listed in Table.~\ref{tab:2dmaze}. Though there are 21 grasps available in total, only one is compatible with the narrow passages in the maze. The proposed framework shows superiority in both the efficiency and success rate, while the naive bidirectional RRT-based methods lack the ability to find a feasible motion plan in the given time budget. Due to the inherent probabilistic completeness, we also observe an increasing success rate with increasing number of threads. 

% Table

\subsection{3D Pick and Place Evaluation}
% \begin{figure}[h!]
%     \centering
%     \includegraphics[width=0.4\textwidth]{Plots_and_Plans/superlinear.png}
%     \caption{Speedup observed for B4P for the Stick-on-Shelf environment}
%     \label{fig:StickSpeed}
% \end{figure}

\begin{figure}[!t]   
    % \centerline{\input{graph/sim_rot.tikz}}
    \centerline{% This file was created with tikzplotlib v0.10.1.
\begin{tikzpicture}

% \definecolor{darkgray176}{RGB}{176,176,176}
% \definecolor{green01270}{RGB}{0,127,0}
% \definecolor{lightgray204}{RGB}{204,204,204}

\definecolor{darkgray176}{rgb}{0.23529,0.72941,0.32941}%
\definecolor{green01270}{rgb}{0.95686,0.76078,0.05098}%
\definecolor{lightgray204}{rgb}{0.85882,0.19608,0.21176}%
\definecolor{mycolor4}{rgb}{0.28235,0.52157,0.92941}%
\definecolor{mycolor5}{rgb}{1.00000,0.54902,0.00000}%

\begin{axis}[
width=0.99\columnwidth,
height=0.5\columnwidth,
legend cell align={left},
legend style={
  fill opacity=0.8,
  draw opacity=1,
  text opacity=1,
  at={(0.03,0.97)},
  anchor=north west,
  draw=none
},
% tick align=outside,
% tick pos=left,
% title={Speedup vs. Number of Threads},
x grid style={darkgray176},
xlabel={\#Threads},
xmajorgrids,
xmin=-0.45, xmax=31.45,
% xtick style={color=black},
% y grid style={darkgray176},
ylabel={Speedup},
ymajorgrids,
ymin=-3.186453202, ymax=88.915517242,
ytick style={color=black},
axis x line*=bottom,
axis y line*=left
]
\addplot [semithick, blue, mark=*, mark size=2, mark options={solid}]
table {%
1 1
2 3.453815261
4 7.110670139
8 20.97560976
16 36.90987124
30 84.72906404
};
\addlegendentry{Observed Speedup}
\addplot [semithick, green01270, mark=*, mark size=2, mark options={solid}]
table {%
1 1
2 2
4 4
8 8
16 16
30 30
};
\addlegendentry{Linear Speedup}
\end{axis}

\end{tikzpicture}}
    \caption{Speedup observed for B4P for the Stick-on-Shelf task}
    \label{fig:StickSpeed}
    \vspace{-0.4cm}
\end{figure}
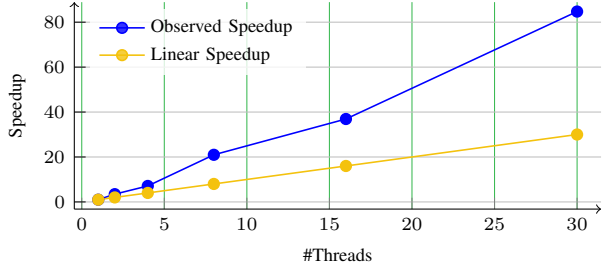

\begin{table}[h]
\centering
\setlength{\tabcolsep}{4pt} % Default value: 6pt

\begin{tabular}{|c|c|c|c|c|c|c|}

\hline
\#Threads & \multicolumn{2}{c|}{Shared} & \multicolumn{2}{c|}{Individual} & \multicolumn{2}{c|}{B4P (Ours)} \\
 & Time & Rate & Time & Rate & Time & Rate \\
\hline
1 & - & 0/10 & - & 0/10 & 172 & 1/10 \\
2 & - & 0/10 & - & 0/10 & 49.8 & 2/10 \\
4 & - & 0/10 & - & 0/10 & 24.19 $\pm 10.62$ & 6/10 \\
8 & - & 0/10 & - & 0/10 & 8.20 $\pm$8.54 & 10/10 \\
16 & - & 0/10 & - & 0/10 & 4.66 $\pm$ 2.33& 10/10 \\
30 & - & 0/10 & 103 & 2/10 & 2.03 $\pm$ 1.45 & 10/10 \\
\hline
\end{tabular}
\caption{Computation time (in seconds) for each method on the Stick-on-Shelf Task to output a feasible trajectory plan. The time budget is 180 seconds.
\label{tab2}
}
\end{table}
% \begin{table}[h]
%     \centering
%     \begin{tabular}{|c|c|c|c|}
%         \hline
%         Threads & Individual & Shared & B4P \\
%         \hline
%         1  & N/A & N/A  & 172 \\
%         2  & N/A  & N/A  & 49.8  \\
%         4  & N/A  & N/A  & 24.19  \\
%         8  & N/A  & N/A & 8.20 \\
%         16  & N/A  & N/A  & 4.66  \\
%         30  & N/A  & 103  & 2.03  \\
%         \hline
%     \end{tabular}
%     \caption{Solve times for the 3D StickShelf problem with a time budget of 180 seconds}
%     \label{tab:stickshelf}
% \end{table}

To further demonstrate our approach's effectiveness, we evaluate it with more realistic and challenging task environments in 3D.  \textcolor{rev}{Our results for the speedup of B4P are compared against using B4P with only one thread.} 
\subsubsection{Stick on Shelf Task}
In this task, the manipulator needs first to pick the yellow stick and place it in the green goal region on the blue shelf (as shown in Fig.~\ref{fig:exps_tasks}-\textit{Stick on Shelf}). The evaluated performance is demonstrated in Table.~\ref{tab2}. As the dimension increases, the naive RRT-based baselines mostly fail to deliver a feasible plan in such a short time budget. Due to the relaxed formulation of tree connections and the path repair design, the proposed method balances the result completeness and efficiency. The proposed B4P possesses a substantial performance increase compared to the two baseline methods in accuracy and speed, along with a superlinear speedup as visualized in Fig.~\ref{fig:StickSpeed}. 

\subsubsection{Grocery Shelf Task}
\begin{table}[h]
\setlength{\tabcolsep}{4pt} % Default value: 6pt

\label{tab:stickshelf}
\centering
\begin{tabular}{|c|c|c|c|c|c|c|}
\hline
\#Threads & \multicolumn{2}{c|}{Shared} & \multicolumn{2}{c|}{Individual} & \multicolumn{2}{c|}{B4P (Ours)} \\
 & Time & Rate & Time & Rate & Time & Rate \\
\hline
1 & - & 0/10 & - & 0/10 & 4.50 $\pm$ 6.38 & 10/10 \\
2 & - & 0/10 & - & 0/10 & 1.93 $\pm$ 1.03 & 10/10 \\
4 & - & 0/10 & - & 0/10 & 0.924 $\pm$ 0.69 & 10/10 \\
8 & - & 0/10 & 28.52 & 2/10 & 0.484 $\pm$ 0.253 & 10/10 \\
16 & - & 0/10 & 27.4 $\pm$ 37.3 & 3/10 & 0.237 $\pm$ 0.13 & 10/10 \\
30 & - & 0/10 & 26.2 $\pm$ 24.7 & 5/10 & 0.124 $\pm$ 0.034 & 10/10 \\
\hline
\end{tabular}
\caption{Computation time (in seconds) for each method on the Grocery-Shelf Task to output a feasible trajectory plan. The time budget is 120 seconds.
\label{tab3}
}
\end{table}

% \begin{table}[h]
%     \centering
%     \begin{tabular}{|c|c|c|c|}
%         \hline
%         Threads & Individual & Shared & B4P \\
%         \hline
%         1  & N/A & N/A  & 4.50\\
%         2  & N/A  & N/A  & 1.93  \\
%         4  & N/A  & N/A  & 0.924  \\
%         8  & N/A  & 28.52 & 0.484 \\
%         16  & N/A  &  27.39 & 0.237  \\
%         30  & N/A  & 26.21  & 0.124 \\
%         \hline
%     \end{tabular}
%     \caption{Solve times for the 3D GroceryShelf problem with a time budget of 120 seconds}
%     \label{tab:example}
% \end{table}
% \begin{figure}[h!]
%     \centering
%     \includegraphics[width=0.4\textwidth]{Plots_and_Plans/YCBActuallySuperLinear.png}
%     \caption{Speedup observed for B4P for the Grocery-Shelf environment}
%     \label{fig:YCBSuperLinear}
% \end{figure}

\begin{figure}[!t]   
    % \centerline{\input{graph/sim_rot.tikz}}
    \centerline{% This file was created with tikzplotlib v0.10.1.
\begin{tikzpicture}

% \definecolor{darkgray176}{RGB}{176,176,176}
% \definecolor{green01270}{RGB}{0,127,0}
% \definecolor{lightgray204}{RGB}{204,204,204}

\definecolor{darkgray176}{rgb}{0.23529,0.72941,0.32941}%
\definecolor{green01270}{rgb}{0.95686,0.76078,0.05098}%
\definecolor{lightgray204}{rgb}{0.85882,0.19608,0.21176}%
\definecolor{mycolor4}{rgb}{0.28235,0.52157,0.92941}%
\definecolor{mycolor5}{rgb}{1.00000,0.54902,0.00000}%

\begin{axis}[
width=0.99\columnwidth,
height=0.5\columnwidth,
legend cell align={left},
legend style={
  fill opacity=0.8,
  draw opacity=1,
  text opacity=1,
  at={(0.03,0.97)},
  anchor=north west,
  draw=none
},
% tick align=outside,
% tick pos=left,
% title={Speedup vs. Number of Threads},
x grid style={darkgray176},
xlabel={\#Threads},
xmajorgrids,
xmin=-0.45, xmax=31.45,
% xtick style={color=black},
% y grid style={darkgray176},
ylabel={Speedup},
ymajorgrids,
ymin=-3.186453202, ymax=88.915517242,
ytick style={color=black},
axis x line*=bottom,
axis y line*=left
]
\addplot [semithick, blue, mark=*, mark size=2, mark options={solid}]
table {%
1 1
2 2.337662338
4 4.87012987
8 9.307135471
16 18.98734177
30 36.42928267
};
\addlegendentry{Observed Speedup}
\addplot [semithick, green01270, mark=*, mark size=2, mark options={solid}]
table {%
1 1
2 2
4 4
8 8
16 16
30 30
};
\addlegendentry{Linear Speedup}
\end{axis}

\end{tikzpicture}}
    \caption{Speedup observed for B4P for the Grocery-Shelf task}
    \label{pic:sim_graph}
    \label{fig:YCBSuperLinear}
\end{figure}
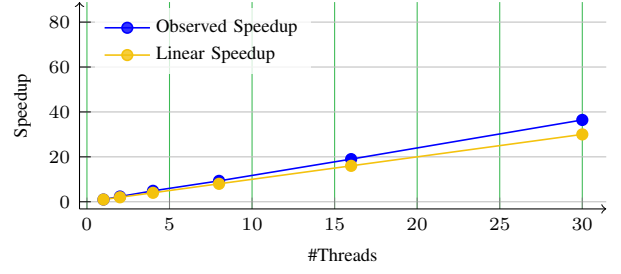

In this task, the target object, the can of spam, must be placed in the goal region in an upward orientation. Compared with the stick-on-shelf problem, additional obstacles are added to the environment to simulate the challenging narrow passage. Three different grasp pose examples are visualized in the right figure from the bottom row of Fig.~\ref{fig:exps_tasks}. Such an intricate task environment creates a narrow path in the configuration space for the pick and place task, only the horizontal grasp pose can satisfy a placement configuration simultaneously with a feasible motion trajectory as displayed in the solution path in Fig.~\ref{fig:exps_tasks}. The experimental result is listed in Table.~\ref{tab3}. As the proposed B4P framework expands unbiased different trees that can fully explore the configuration space, it outperforms both baselines by a large margin and guarantees a feasible trajectory plan in the given time budget. Meanwhile, we also observe a superlinear speed when increasing the threads for computing as shown in Fig.~\ref{fig:YCBSuperLinear}. This feature possesses an inherent advantage for accelerating scalable applications in the real world. \textcolor{rev}{The speedup observed for this task is lower than that of the Stick-on-Shelf task, likely due to the greater number of valid grasps available for this task.} 

% \begin{figure}[h!]
%     \centering
%     \includegraphics[width=0.5\textwidth]{Plots_and_Plans/GShelfOverview.pdf}
%     \caption{Pictured on the left is the GroceryShelf problem. The robot must relocate the spam, to the right of all the obstacles, to the small green region behind the other objects in the red shelf. Pictured on the right are 4 sampled grasps of the target object. }
%     \label{fig:spam_shelf}
% \end{figure}
% \begin{figure}[h!]
%     \centering
%     \includegraphics[width=0.5\textwidth]{Plots_and_Plans/StickShelf.png}
%     \caption{StickShelf Environment}
%     \label{fig:Stick_shelf}
% \end{figure}

\section{Conclusion}
In this work, we proposed B4P, a framework to address the problem of simultaneously finding grasp pose and motion plan for downstream placement tasks. By leveraging the parallelized bidirectional forests with path repair, the proposed framework demonstrates significant efficiency together with result completeness on diverse scenarios. Additionally, we investigated its inherent parallelism to achieve a superlinear speedup.

We plan to extend the current work in the following directions: adding perception modules to get rid of the assumption of a perfect description of the environment and extending a re-grasping policy to tackle extreme situations that no compatible grasp poses in the initial object state.

\bibliographystyle{IEEEtran}
\bibliography{references}

\begin{thebibliography}{10}
\providecommand{\url}[1]{#1}
\csname url@rmstyle\endcsname
\providecommand{\newblock}{\relax}
\providecommand{\bibinfo}[2]{#2}
\providecommand\BIBentrySTDinterwordspacing{\spaceskip=0pt\relax}
\providecommand\BIBentryALTinterwordstretchfactor{4}
\providecommand\BIBentryALTinterwordspacing{\spaceskip=\fontdimen2\font plus
\BIBentryALTinterwordstretchfactor\fontdimen3\font minus \fontdimen4\font\relax}
\providecommand\BIBforeignlanguage[2]{{%
\expandafter\ifx\csname l@#1\endcsname\relax
\typeout{** WARNING: IEEEtran.bst: No hyphenation pattern has been}%
\typeout{** loaded for the language `#1'. Using the pattern for}%
\typeout{** the default language instead.}%
\else
\language=\csname l@#1\endcsname
\fi
#2}}

\bibitem{correll2016analysis}
N.~Correll, K.~E. Bekris, D.~Berenson, O.~Brock, A.~Causo, K.~Hauser, K.~Okada, A.~Rodriguez, J.~M. Romano, and P.~R. Wurman, ``Analysis and observations from the first amazon picking challenge,'' \emph{IEEE Transactions on Automation Science and Engineering}, vol.~15, no.~1, pp. 172--188, 2016.

\bibitem{wisspeintner2009robocup}
T.~Wisspeintner, T.~Van Der~Zant, L.~Iocchi, and S.~Schiffer, ``Robocup@ home: Scientific competition and benchmarking for domestic service robots,'' \emph{Interaction Studies}, vol.~10, no.~3, pp. 392--426, 2009.

\bibitem{morrison2018closing}
D.~Morrison, P.~Corke, and J.~Leitner, ``Closing the loop for robotic grasping: A real-time, generative grasp synthesis approach,'' \emph{Robotics: Science and Systems XIV}, pp. 1--10, 2018.

\bibitem{fang2020graspnet}
H.-S. Fang, C.~Wang, M.~Gou, and C.~Lu, ``Graspnet-1billion: A large-scale benchmark for general object grasping,'' in \emph{Proceedings of the IEEE/CVF conference on computer vision and pattern recognition}, 2020, pp. 11\,444--11\,453.

\bibitem{hang2016hierarchical}
K.~Hang, M.~Li, J.~A. Stork, Y.~Bekiroglu, F.~T. Pokorny, A.~Billard, and D.~Kragic, ``Hierarchical fingertip space: A unified framework for grasp planning and in-hand grasp adaptation,'' \emph{IEEE Transactions on robotics}, vol.~32, no.~4, pp. 960--972, 2016.

\bibitem{Elbanhawi2014}
M.~Elbanhawi and M.~Simic, ``Sampling-based robot motion planning: A review,'' \emph{IEEE Access}, vol.~2, pp. 56--77, 2014.

\bibitem{Harada12}
K.~Harada, T.~Tsuji, K.~Nagata, N.~Yamanobe, H.~Onda, T.~Yoshimi, and Y.~Kawai, ``Object placement planner for robotic pick and place tasks,'' in \emph{IEEE/RSJ International Conference on Intelligent Robots and Systems}, 2012, pp. 980--985.

\bibitem{haustein2019object}
J.~A. Haustein, K.~Hang, J.~Stork, and D.~Kragic, ``Object placement planning and optimization for robot manipulators,'' in \emph{2019 IEEE/RSJ International Conference on Intelligent Robots and Systems (IROS)}.\hskip 1em plus 0.5em minus 0.4em\relax IEEE, 2019, pp. 7417--7424.

\bibitem{GraspPlanningHaustein}
J.~A. Haustein, K.~Hang, and D.~Kragic, ``Integrating motion and hierarchical fingertip grasp planning,'' in \emph{2017 IEEE International Conference on Robotics and Automation (ICRA)}, 2017, pp. 3439--3446.

\bibitem{plaku2005sampling}
E.~Plaku, K.~E. Bekris, B.~Y. Chen, A.~M. Ladd, and L.~E. Kavraki, ``Sampling-based roadmap of trees for parallel motion planning,'' \emph{IEEE Transactions on Robotics}, vol.~21, no.~4, pp. 597--608, 2005.

\bibitem{bicchi2000robotic}
A.~Bicchi and V.~Kumar, ``Robotic grasping and contact: A review,'' in \emph{Proceedings 2000 ICRA. Millennium conference. IEEE international conference on robotics and automation. Symposia proceedings (Cat. No. 00CH37065)}, vol.~1.\hskip 1em plus 0.5em minus 0.4em\relax IEEE, 2000, pp. 348--353.

\bibitem{newbury2023deep}
R.~Newbury, M.~Gu, L.~Chumbley, A.~Mousavian, C.~Eppner, J.~Leitner, J.~Bohg, A.~Morales, T.~Asfour, D.~Kragic, \emph{et~al.}, ``Deep learning approaches to grasp synthesis: A review,'' \emph{IEEE Transactions on Robotics}, vol.~39, no.~5, pp. 3994--4015, 2023.

\bibitem{ciocarlie2009hand}
M.~T. Ciocarlie and P.~K. Allen, ``Hand posture subspaces for dexterous robotic grasping,'' \emph{The International Journal of Robotics Research}, vol.~28, no.~7, pp. 851--867, 2009.

\bibitem{liu2021synthesizing}
T.~Liu, Z.~Liu, Z.~Jiao, Y.~Zhu, and S.-C. Zhu, ``Synthesizing diverse and physically stable grasps with arbitrary hand structures using differentiable force closure estimator,'' \emph{IEEE Robotics and Automation Letters}, vol.~7, no.~1, pp. 470--477, 2021.

\bibitem{chen2023differentiable}
Y.~Chen, X.~Gao, K.~Yao, L.~Niederhauser, Y.~Bekiroglu, and A.~Billard, ``Differentiable robot neural distance function for adaptive grasp synthesis on a unified robotic arm-hand system,'' \emph{arXiv preprint arXiv:2309.16085}, 2023.

\bibitem{sundermeyer2021contact}
M.~Sundermeyer, A.~Mousavian, R.~Triebel, and D.~Fox, ``Contact-graspnet: Efficient 6-dof grasp generation in cluttered scenes,'' in \emph{2021 IEEE International Conference on Robotics and Automation (ICRA)}.\hskip 1em plus 0.5em minus 0.4em\relax IEEE, 2021, pp. 13\,438--13\,444.

\bibitem{orthey2023sampling}
A.~Orthey, C.~Chamzas, and L.~E. Kavraki, ``Sampling-based motion planning: A comparative review,'' \emph{Annual Review of Control, Robotics, and Autonomous Systems}, vol.~7, 2023.

\bibitem{thomason2024motions}
W.~Thomason, Z.~Kingston, and L.~E. Kavraki, ``Motions in microseconds via vectorized sampling-based planning,'' in \emph{2024 IEEE International Conference on Robotics and Automation (ICRA)}.\hskip 1em plus 0.5em minus 0.4em\relax IEEE, 2024, pp. 8749--8756.

\bibitem{bialkowski2011massively}
J.~Bialkowski, S.~Karaman, and E.~Frazzoli, ``Massively parallelizing the rrt and the rrt,'' in \emph{2011 IEEE/RSJ International Conference on Intelligent Robots and Systems}.\hskip 1em plus 0.5em minus 0.4em\relax IEEE, 2011, pp. 3513--3518.

\bibitem{fishman2023motion}
A.~Fishman, A.~Murali, C.~Eppner, B.~Peele, B.~Boots, and D.~Fox, ``Motion policy networks,'' in \emph{conference on Robot Learning}.\hskip 1em plus 0.5em minus 0.4em\relax PMLR, 2023, pp. 967--977.

\bibitem{sundaralingam2023curobo}
B.~Sundaralingam, S.~K.~S. Hari, A.~Fishman, C.~Garrett, K.~Van~Wyk, V.~Blukis, A.~Millane, H.~Oleynikova, A.~Handa, F.~Ramos, \emph{et~al.}, ``Curobo: Parallelized collision-free robot motion generation,'' in \emph{2023 IEEE International Conference on Robotics and Automation (ICRA)}.\hskip 1em plus 0.5em minus 0.4em\relax IEEE, 2023, pp. 8112--8119.

\bibitem{bhardwaj2022storm}
M.~Bhardwaj, B.~Sundaralingam, A.~Mousavian, N.~D. Ratliff, D.~Fox, F.~Ramos, and B.~Boots, ``Storm: An integrated framework for fast joint-space model-predictive control for reactive manipulation,'' in \emph{Conference on Robot Learning}.\hskip 1em plus 0.5em minus 0.4em\relax PMLR, 2022, pp. 750--759.

\bibitem{kuffner2000rrt}
J.~J. Kuffner and S.~M. LaValle, ``Rrt-connect: An efficient approach to single-query path planning,'' in \emph{Proceedings 2000 ICRA. Millennium conference. IEEE international conference on robotics and automation. Symposia proceedings (Cat. No. 00CH37065)}, vol.~2.\hskip 1em plus 0.5em minus 0.4em\relax IEEE, 2000, pp. 995--1001.

\bibitem{lindemann2004incrementally}
S.~R. Lindemann and S.~M. LaValle, ``Incrementally reducing dispersion by increasing voronoi bias in rrts,'' in \emph{IEEE International Conference on Robotics and Automation, 2004. Proceedings. ICRA'04. 2004}, vol.~4.\hskip 1em plus 0.5em minus 0.4em\relax IEEE, 2004, pp. 3251--3257.

\bibitem{zeng2022robotic}
A.~Zeng, S.~Song, K.-T. Yu, E.~Donlon, F.~R. Hogan, M.~Bauza, D.~Ma, O.~Taylor, M.~Liu, E.~Romo, \emph{et~al.}, ``Robotic pick-and-place of novel objects in clutter with multi-affordance grasping and cross-domain image matching,'' \emph{The International Journal of Robotics Research}, vol.~41, no.~7, pp. 690--705, 2022.

\bibitem{saut2010planning}
J.-P. Saut, M.~Gharbi, J.~Cort{\'e}s, D.~Sidobre, and T.~Sim{\'e}on, ``Planning pick-and-place tasks with two-hand regrasping,'' in \emph{2010 IEEE/RSJ International Conference on Intelligent Robots and Systems}.\hskip 1em plus 0.5em minus 0.4em\relax IEEE, 2010, pp. 4528--4533.

\bibitem{shanthi2024pick}
M.~D. Shanthi and T.~Hermans, ``Pick and place planning is better than pick planning then place planning,'' \emph{IEEE Robotics and Automation Letters}, vol.~9, no.~3, pp. 2790--2797, 2024.

\bibitem{he2023pick2place}
Z.~He, N.~Chavan-Dafle, J.~Huh, S.~Song, and V.~Isler, ``Pick2place: Task-aware 6dof grasp estimation via object-centric perspective affordance,'' in \emph{2023 IEEE International Conference on Robotics and Automation (ICRA)}.\hskip 1em plus 0.5em minus 0.4em\relax IEEE, 2023, pp. 7996--8002.

\bibitem{brooks1983planning}
R.~A. Brooks, ``Planning collision-free motions for pick-and-place operations,'' \emph{The International Journal of Robotics Research}, vol.~2, no.~4, pp. 19--44, 1983.

\bibitem{mujoco}
E.~Todorov, T.~Erez, and Y.~Tassa, ``Mujoco: A physics engine for model-based control,'' in \emph{2012 IEEE/RSJ International Conference on Intelligent Robots and Systems}, 2012, pp. 5026--5033.

\end{thebibliography}
\end{document}